\documentclass{article}

\PassOptionsToPackage{numbers,sort&compress}{natbib}
\usepackage[dblblindworkshop,final]{neurips_2026}  % camera-ready
\workshoptitle{Who Verifies the Agents? Toward Reliable Agent Development}

\usepackage[utf8]{inputenc}
\usepackage[T1]{fontenc}
\usepackage{amsmath,amssymb,amsthm}
\usepackage{mathtools}
\usepackage{bm}
\usepackage{graphicx}
\usepackage{booktabs}
\usepackage{multirow}
\usepackage{array}
\usepackage{xcolor}
\usepackage{algorithm}
\usepackage{algpseudocode}
\usepackage{enumitem}
\usepackage{mdframed}
\usepackage{makecell}
\usepackage{tikz}
\usepackage{pgfplots}
\pgfplotsset{compat=1.17}
\usepackage{mathrsfs}
\usepackage[hidelinks]{hyperref}
\usetikzlibrary{arrows.meta,positioning,shapes.geometric,decorations.pathreplacing}

\newtheorem{definition}{Definition}[section]

\newcommand{\R}{\mathbb{R}}
\newcommand{\E}{\mathbb{E}}
\newcommand{\Prob}{\mathbb{P}}
\newcommand{\calS}{\mathcal{S}}
\newcommand{\calA}{\mathcal{A}}
\newcommand{\calJ}{\mathcal{J}}
\newcommand{\calD}{\mathcal{D}}
\newcommand{\bW}{\bm{W}}
\newcommand{\btheta}{\bm{\theta}}
\newcommand{\blambda}{\bm{\lambda}}
\newcommand{\dkl}{D_{\mathrm{KL}}}

\title{VACS: Value-Aligned Compositional Shielding for Multi-Agent Reasoning}

\author{
Yiyao Zhang$^{a}$,
Diksha Goel$^{b}$,
Hussain Ahmad$^{a}$\\[0.8em]
\begin{minipage}{0.92\textwidth}
\centering\small
$^{a}$School of Computer Science and Information Technology, Adelaide University, Adelaide, SA, Australia\\
$^{b}$CSIRO's Data61, Clayton, VIC, Australia\\[0.4em]
\textit{Author e-mail addresses:}
yiyao.zhang@student.adelaide.edu.au (Y. Zhang),
diksha.goel@csiro.au (D. Goel),
hussain.ahmad@adelaide.edu.au (H. Ahmad)
\end{minipage}
}

\begin{document}

\maketitle

\begin{abstract}
Multi-agent reasoning systems in high-stakes domains must be both accurate and safe, yet agents often follow heterogeneous value priorities (e.g., rigor, conciseness, safety), causing conflicting recommendations. Existing methods do not jointly provide: (i) principled inference of each agent's implicit values from behavior, (ii) compositional formal safety guarantees without full online communication, and (iii) value-aware conflict resolution with faithful explanations.
We present \textbf{VACS} (\textbf{V}alue-\textbf{A}ligned \textbf{C}ompositional \textbf{S}hielding), a four-layer framework addressing all three. \textbf{Layer 1} learns value-dimension rewards from pairwise preferences using Bradley--Terry modeling and infers per-agent value weights via deep MaxEnt IRL. \textbf{Layer 2} encodes value constraints in a Lean-inspired DSL and synthesizes compositional assume-guarantee shields for runtime safety. \textbf{Layer 3} resolves disagreement through nucleolus-based credit allocation and Hamiltonian consensus optimization under long-term value constraints. \textbf{Layer 4} extracts a critical reasoning path from co-state sensitivities and generates formally grounded natural-language explanations.
Our contribution is primarily a unified systems design with formalized interfaces and operational guarantees at the verifier-constrained decision level, rather than a complete end-to-end formal proof of all language-model internals.
In controlled proof-of-concept evaluations with role-conditioned agent panels on NEJM-AI QA, MathInstruct-Subset, and a cybersecurity incident-response benchmark (CyberSec-Eval), VACS outperforms strong baselines in accuracy (85.4\%, 95.0\%, and 90.0\%) while reducing logical inconsistency rates to near zero.
\end{abstract}

\section{Introduction}
\label{sec:intro}

Multi-agent reasoning is a powerful paradigm for complex decision-making in high-stakes domains such as theorem proving and clinical diagnostics: pooling diverse trajectories from agents with distinct skills and perspectives consistently improves factuality, robustness, and task performance \citep{du2024improving}.

Despite this potential, collaborative multi-agent setups introduce a fundamental challenge: individual agents almost always operate under heterogeneous, implicit value priorities, in proof verification, strict step-by-step rigor versus intuitive conciseness; in clinical reasoning, maximum diagnostic coverage versus cost-consciousness and patient risk.
In current practice, multi-agent systems typically aggregate outputs using simple, ad-hoc rules such as majority voting or confidence-based weighting. These strategies merely pool final outputs and fail to reconcile the underlying value conflicts between agents \citep{informed2025voting,li2025nucleolus}, so the final group decision can easily become internally inconsistent, unsafe, or impossible to faithfully justify. There is a pressing need for a framework that can jointly: (i) infer each agent's implicit value system from past behavior \citep{ziebart2008maximum}, (ii) formally guarantee that collective reasoning steps adhere to safe boundaries \citep{brorholt2025compositional}, and (iii) resolve inter-agent disagreements through principled, value-aware consensus rather than naive statistical aggregation.

To bridge this gap, we propose \textbf{VACS} (\textbf{V}alue-\textbf{A}ligned \textbf{C}ompositional \textbf{S}hielding), a unified, four-layer framework that integrates value inference, formal safety, and cooperative consensus into multi-agent reasoning through a top-to-bottom pipeline: Bradley--Terry preference modeling and deep Maximum Entropy Inverse Reinforcement Learning (MaxEnt IRL) infer each agent's implicit value profile; the learned values are compiled into formal constraints in a Lean-inspired Domain-Specific Language (DSL) that drive compositional assume-guarantee (AG) shields filtering unsafe actions at runtime; disagreements are resolved as a cooperative game via nucleolus credit allocation and Hamiltonian constrained optimization; and critical reasoning paths extracted from Hamiltonian sensitivities ground natural-language explanations in verified proof records. We frame VACS as an integrative framework that operationalizes existing IRL, verification, and cooperative-optimization ideas under one deployable pipeline. Our core contributions are:
\begin{itemize}[leftmargin=*,label=\textbullet,itemsep=1pt,topsep=2pt]
  \item \textbf{Implicit value discovery via IRL:} agent value systems are treated as latent variables inferred from behavioral trajectories via a joint Bradley--Terry and deep MaxEnt IRL pipeline.
  \item \textbf{Compositional safety guarantees:} learned values become Lean-DSL constraints with verifier-mediated operational guarantees under stated AG assumptions: locally shielded agents guarantee global logical consistency without inter-agent communication during execution.
  \item \textbf{Economics-grounded conflict resolution:} nucleolus-based credit allocation and Hamiltonian optimization produce a value-weighted agreement that is stable and fair to all coalitions.
  \item \textbf{Auditable explanation validation:} a hybrid structural--gradient explanation layer is validated for decision-level faithfulness via expert judgment and intervention-based tests.
  \item \textbf{Empirical results:} on MathInstruct-Subset, NEJM-AI QA, and CyberSec-Eval, VACS outperforms strong baselines in accuracy (up to 95.0\%) while reducing logical inconsistency to zero.
\end{itemize}

\section{Related Work}
\label{sec:related}

VACS builds on four mature lines of work: multi-agent/ensemble reasoning, inverse reinforcement learning, formal verification, and cooperative consensus. Multi-agent LLM systems improve robustness and factuality through diversity, but usually rely on heuristic aggregation rather than explicit modeling of heterogeneous values \citep{du2024improving,openai2023gpt4,touvron2023llama2,multiagent2024survey}; role-specialised LLM panels that score, propose, and select under a shared objective make the aggregation step particularly load-bearing \citep{chen2025trader}. Multi-agent reinforcement learning for autonomous cyber defence raises the same pairing of coordination and auditability that motivates our Layers~3--4 \citep{zhang2026explainable,goel2025coevolutionary}. IRL and preference-learning methods provide a principled way to recover latent objectives from behavior and pairwise comparisons \citep{ng2000algorithms,ziebart2008maximum,wulfmeier2015maximum,christiano2017deep}, while formal verification and compositional shielding offer deployable safety guarantees for structured reasoning pipelines \citep{katz2017reluplex,gehr2018ai2,brorholt2025compositional,moura2021lean4}; a complementary line gates \emph{when} an agent should act or escalate rather than constraining \emph{what} it may execute \citep{zhang2026meta}. Cooperative-game and voting-based aggregation methods study stable consensus formation, but typically do not incorporate verifier-backed value constraints \citep{li2025nucleolus,informed2025voting}. VACS integrates these complementary capabilities, value profiles from preferences (Layer~1), compositional safety constraints (Layer~2), coalition-aware conflict resolution (Layer~3), and verifier-tied decision-level explanations (Layer~4), rather than replacing any one paradigm; this combination is the framework's main distinction.

\section{Preliminaries and Notation}
\label{sec:prelim}

\paragraph{Multi-agent reasoning setting.}
We consider a panel of $n$ agents $\calJ = \{1, \ldots, n\}$, where each agent $j \in \calJ$ may be a human specialist or an AI model. Given a reasoning task $\tau$ (e.g., a mathematical problem or clinical scenario), each agent $j$ produces a reasoning trajectory $\xi^j = (s_0^j, a_0^j, s_1^j, a_1^j, \ldots, s_T^j)$, where $s_t^j$ is the agent's reasoning state at step $t$ and $a_t^j$ is its reasoning action (e.g., applying a proof rule, citing a precedent). The trajectory terminates with a final answer $y^j \in \calA$.

\begin{definition}[Value Dimension and Value Profile]
  A \emph{value dimension} $v_k$ is a property of reasoning trajectories that agents may care about to varying degrees, associated with a reward function $R_{v_k}: \calS \times \calA \rightarrow \R$. We consider $m=5$ dimensions: logical completeness ($v_1$), conciseness ($v_2$), generalisability ($v_3$), logical soundness ($v_4$), and safety/constraint satisfaction ($v_5$). Agent $j$'s \emph{value profile} is a weight vector $\bW^j = (W^j_1, \ldots, W^j_m) \in \Delta^{m-1}$, where $W^j_k$ is the relative importance agent $j$ places on $v_k$; the agent's composite reward is $R^j(\xi) = \sum_{k=1}^m W^j_k \cdot R_{v_k}(\xi)$.
\end{definition}

\paragraph{Operationalization of ``value'' as preference/reward dimensions.}
``Value'' is operationalized as a \emph{preference/reward dimension} inferred from comparative judgments: each $v_k$ is an empirically measurable criterion over trajectories, with definitions fixed before training (e.g., logical completeness = fraction of required sub-claims addressed; safety = proportion of steps satisfying domain safety constraints). Throughout, ``implicit value system'' is shorthand for the latent preference profile over reward dimensions estimated by Bradley--Terry + MaxEnt IRL. Annotation sources, protocol, reliability statistics, and profile-stability checks are given in Appendix~\ref{app:operationalization}.

\paragraph{Lean-DSL constraint language.}
To make agent reasoning verifiable, we represent each value requirement as a formal rule in a Lean-inspired DSL. Intuitively, each rule says: \emph{if certain conditions hold before an action, then a required condition must hold after the action}. Formally, for value dimension $v_k$, we write a constraint $\phi_k$ as
\begin{equation}
  \phi_k \equiv \forall s \in \calS,\; a \in \calA:\; \Psi_k(s, a) \Rightarrow \Omega_k(s'),
  \label{eq:lean_constraint}
\end{equation}
where $\Psi_k(s,a)$ is the \textbf{precondition} (what must hold before taking action $a$ in state $s$) and $\Omega_k(s')$ is the \textbf{postcondition} (what must hold in the next state $s'$). For example, for logical completeness ($\phi_1$), when an agent applies a proof step, all required premises for that step must be verified in the resulting state.

\section{The VACS Framework}
\label{sec:framework}

\begin{figure}[t]
  \centering
  \includegraphics[width=0.82\textwidth]{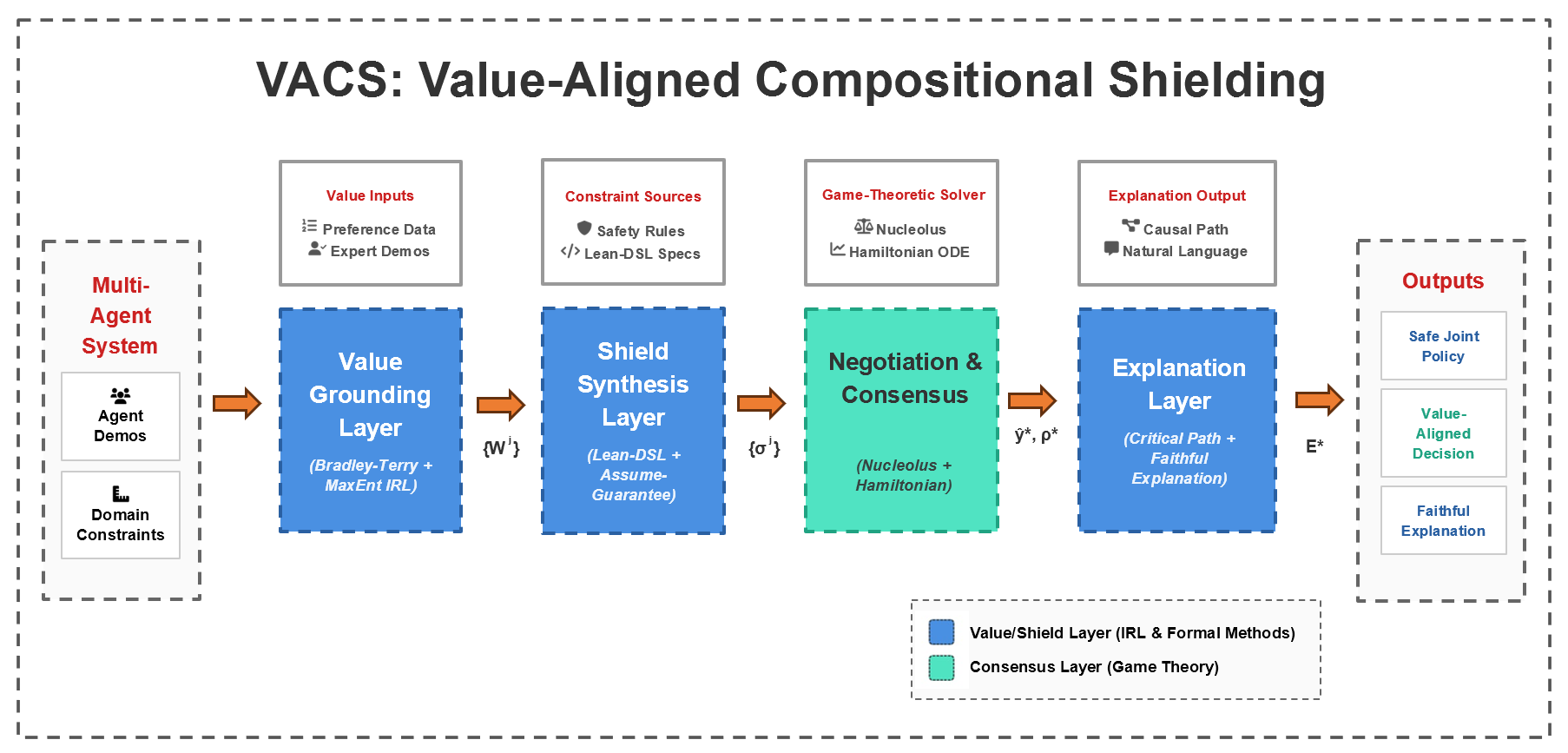}
  \caption{End-to-end VACS architecture for one query: Layer~1 infers per-agent value profiles, Layer~2 enforces compositional safety constraints, Layer~3 performs value-aware consensus, and Layer~4 produces an auditable explanation record.}
  \label{fig:vacs_architecture}
\end{figure}

VACS is organised into four layers with a clean top-to-bottom information flow (Figure~\ref{fig:vacs_architecture}): Layer~1 produces per-agent value profiles $\{\bW^j\}$ consumed by Layers~2 and~3; Layer~2 produces per-agent shields $\{\sigma^j\}$ that wrap agent policies at execution time; Layer~3 produces a consensus answer $\hat{y}^*$ and value alignment score $\rho^*$; and Layer~4 delivers an auditable, decision-relevant explanation $E^*$ to the end user.

\subsection{Layer 1: Multi-Dimensional Value Grounding and Agent Profiling}
\label{sec:layer1}

Layer~1 is a two-stage pipeline: first grounding preference/reward dimensions from comparative data, then inferring each agent's latent preference profile over these dimensions from behavioural trajectories.

\paragraph{Stage 1: Bradley--Terry value function learning.}
For each value dimension $v_k$, we collect a dataset of pairwise comparisons $\calD_k = \{(\xi_i, \xi_j, y_{ij})\}$, where $y_{ij} = 1$ if trajectory $\xi_i$ is preferred over $\xi_j$ on dimension $v_k$ and $y_{ij} = 0$ otherwise. We model the preference probability using the Bradley--Terry model \citep{bradley1952rank}:
\begin{equation}
  \Prob(\xi_i \succ_k \xi_j) = \frac{\exp(R_{v_k}(\xi_i))}{\exp(R_{v_k}(\xi_i)) + \exp(R_{v_k}(\xi_j))},
  \label{eq:bradley_terry}
\end{equation}
and learn $R_{v_k}$ by maximising the log-likelihood over $\calD_k$:
\begin{equation}
  \hat{R}_{v_k} = \arg\max_{R_{v_k}} \sum_{(\xi_i, \xi_j, y_{ij}) \in \calD_k}
  \left[y_{ij} \log \Prob(\xi_i \succ_k \xi_j) + (1-y_{ij}) \log \Prob(\xi_j \succ_k \xi_i)\right].
  \label{eq:bt_mle}
\end{equation}
The reward function $R_{v_k}$ is parameterised as a neural network $R_{v_k,\btheta_k}$ and optimised via stochastic gradient descent.

\paragraph{Stage 2: Deep MaxEnt IRL for agent value profiling.}
Given the learned value reward functions $\{R_{v_k}\}_{k=1}^m$, we infer each agent $j$'s value weight vector $\bW^j$ from its historical reasoning trajectories $\calD^j = \{\xi^j_1, \ldots, \xi^j_{N_j}\}$ using deep MaxEnt IRL \citep{ziebart2008maximum,wulfmeier2015maximum}:
\begin{equation}
  \hat{\bW}^j = \arg\max_{\bW^j \in \Delta^{m-1}} \frac{1}{N_j} \sum_{i=1}^{N_j}
  \log P_{\bW^j}(\xi^j_i) - \lambda_{\mathrm{IRL}} \|\bW^j\|_2^2,
  \label{eq:maxent_irl}
\end{equation}
where $P_{\bW^j}(\xi) \propto \exp\!\left(\sum_{k=1}^m W^j_k \cdot R_{v_k}(\xi)\right)$ is the maximum entropy trajectory distribution under weight vector $\bW^j$. The gradient of the objective is
\begin{equation}
  \nabla_{\bW^j} \mathcal{L}_{\mathrm{IRL}} = \frac{1}{N_j} \sum_{i=1}^{N_j}
  \bm{f}(\xi^j_i) - \E_{P_{\bW^j}}[\bm{f}(\xi)] - 2\lambda_{\mathrm{IRL}} \bW^j,
  \label{eq:maxent_gradient}
\end{equation}
where $\bm{f}(\xi) = (R_{v_1}(\xi), \ldots, R_{v_m}(\xi))^\top$ is the feature vector of trajectory $\xi$; the expectation is computed via dynamic programming over the reasoning state space.

\paragraph{Computational object for MaxEnt IRL.}
For free-form reasoning, IRL is instantiated over a structured abstraction of text traces: states summarize context, working memory, unresolved sub-questions, and risk flags; actions are typed reasoning operators (e.g., \texttt{deduce}, \texttt{verify}, \texttt{revise}) obtained by segmenting chain-of-thought text into operator-level steps; and dynamic programming runs on a discrete abstraction learned by vector quantization over state embeddings (details in Appendix~\ref{app:irl_object}). Finally, $k$-means clustering of the inferred $\{\hat{\bW}^j\}$ into value-type families (e.g., ``rigour-focused'', ``efficiency-focused'') informs the coalition structure in Layer~3 and shield synthesis in Layer~2.

\subsection{Layer 2: Value-Aligned Compositional Shield Synthesis}
\label{sec:layer2}

Layer~2 encodes each value dimension as a formal Lean-DSL constraint and synthesises per-agent compositional shields via AG reasoning.

\paragraph{Lean-DSL constraint encoding.}
For each value dimension $v_k$ and agent $j$, we encode the value constraint as a Lean-DSL formula $\phi^j_k$ specifying the safety requirement for dimension $v_k$ in agent $j$'s local reasoning scope. The per-agent safety specification is the conjunction of all value constraints weighted by the agent's value profile:
\begin{equation}
  \Phi^j = \bigwedge_{k : W^j_k > \theta_{\mathrm{val}}} \phi^j_k,
  \label{eq:agent_safety_spec}
\end{equation}
where $\theta_{\mathrm{val}}$ filters out value dimensions to which agent $j$ assigns negligible weight, so that each shield enforces only the constraints relevant to its agent's value system, avoiding unnecessary conservatism. The five canonical value constraints require that every proof step has all premises verified in the current state ($\phi^j_1$, completeness); that trajectory length does not exceed a state-dependent bound $L_{\max}(s)$ ($\phi^j_2$, conciseness); that steps do not rely on case-specific assumptions absent from the problem statement ($\phi^j_3$, generalisability); that no step contradicts a previously established fact ($\phi^j_4$, soundness); and that no action violates hard constraints in the task description ($\phi^j_5$, safety).

\paragraph{Compositional shield synthesis via AG reasoning.}
After defining each agent's value-based safety rules $\Phi^j$, we build a verifier-guided text shield $\sigma^j:\calS^j \times \calA^j \rightarrow \calA^j$ for each agent. In text reasoning an action is a structured step $a_t^j=(\mathsf{op},\mathsf{args},\mathsf{claim})$ extracted from the generated rationale; unsafe actions are those violating any Lean-DSL constraint in $\Phi^j$. Rather than ``nearest safe replacement'', we use a prune-and-rerank protocol: generate admissible candidate steps, prune all verifier-failing candidates, then rerank only verifier-approved candidates by semantic proximity and factual support. The shield follows a simple rule:
\begin{equation}
  \sigma^j(s^j, a^j)=
  \begin{cases}
    a^j, & \text{if } (s^j,a^j)\models \Phi^j,\\[4pt]
    \displaystyle\arg\min_{a' \models \Phi^j(s^j)} \|a'-a^j\|_{\calA},
    & \text{otherwise}.
  \end{cases}
  \label{eq:vacs_shield}
\end{equation}
If the proposed action is safe, it is kept unchanged; otherwise the shield generates candidates $\mathcal{C}$, keeps only the verifier-approved subset $\mathcal{C}_{\mathrm{ok}} = \{a' \in \mathcal{C} \mid (s_t^j,a') \models \Phi^j\}$, and selects $\arg\max_{a' \in \mathcal{C}_{\mathrm{ok}}} [\lambda_{\mathrm{sem}}\mathrm{Sim}(a',a_t^j)+\lambda_{\mathrm{fact}}\mathrm{Fact}(a',s_t^j)]$, abstaining if $\mathcal{C}_{\mathrm{ok}}=\emptyset$ (pseudocode in Appendix~\ref{app:shield_alg}).

\noindent\textbf{No-new-error guarantee (operational):} because only verifier-approved candidates in $\mathcal{C}_{\mathrm{ok}}$ can be executed, replacement cannot introduce a violation of $\Phi^j$ that was absent from the approved set; if no approved candidate exists, the shield abstains instead of forcing an unsafe rewrite. We emphasize the scope of this guarantee: it is conditional on the finite candidate set produced at each step, it ensures no approved-set violation is executed, not that a correct action is present in the pool. It provides (i) local control (each agent is corrected independently) and (ii) global consistency (unsafe local actions do not propagate into contradictory collective conclusions).

\paragraph{Assume--guarantee interface for compositionality.}
The synthesis is compositional: each shield is constructed assuming other agents also respect their own shields. For each agent $j$, the \emph{local assumption} $A_j$ states that messages from other agents satisfy declared interface predicates over a common verifier-checked symbol layer $\Sigma_{\mathrm{sh}}$, and the \emph{local guarantee} $G_j$ states that every executed action satisfies $\Phi^j$ and preserves local invariant $I_j$; rounds are lock-step with deterministic tie-breaking (asynchronous updates are outside current guarantees). The compositional proof obligations are
\begin{equation}
(O1)\;\forall j:\; A_j \Rightarrow G_j, \qquad
(O2)\;\bigwedge_{j=1}^{n} G_j \Rightarrow \bigwedge_{j=1}^{n} A_j, \qquad
(O3)\;\bigwedge_{j=1}^{n} G_j \Rightarrow I_{\mathrm{global}},
\label{eq:ag_obligations}
\end{equation}
discharged respectively by local shield verification, interface-compatibility checks over $\Sigma_{\mathrm{sh}}$, and induction over synchronized rounds. Under O1--O3, local shielding composes into global consistency without full online sharing of internal reasoning traces.

\subsection{Layer 3: Nucleolus-Weighted Hamiltonian Consensus}
\label{sec:layer3}

Layer~3 integrates nucleolus-based negotiation with Hamiltonian optimisation to produce a globally optimal, value-aware consensus: the nucleolus determines the credit allocation among agents, and the Hamiltonian optimisation uses these credits as weights in the long-term value alignment objective.

\paragraph{Value alignment score.}
For each agent $j$, we compute a \emph{value alignment score} $\rho^j$ measuring how well the agent's recommendations align with the collective value profile $\bar{\bW} = \frac{1}{n} \sum_{j=1}^n \bW^j$:
\begin{equation}
  \rho^j = \exp\!\left(-\dkl\!\left(\bW^j \,\|\, \bar{\bW}\right)\right) \cdot \mathrm{acc}^j,
  \label{eq:value_alignment_score}
\end{equation}
where $\mathrm{acc}^j \in [0,1]$ is agent $j$'s historical accuracy on the current task type. The score fuses value compatibility and empirical reliability multiplicatively: KL to the panel mean captures dimension-wise mismatch with asymmetric penalty and adapts to panel composition without extra learned parameters, while the multiplicative form enforces a conjunctive gate, influence is high only when both alignment and reliability are high, which is desirable in safety-critical consensus where either failure mode is harmful.

\paragraph{Nucleolus-based credit allocation.}
When agents disagree, we need a fair way to decide how much influence each agent should have. We model this as a cooperative game among agents, where a coalition gets more power if its members strongly support the same answer and have high alignment scores. For any coalition $C \subseteq \calJ$:
\begin{equation}
  v(C) = \max_{y \in \calA} \sum_{j \in C} \rho^j \cdot \mathbb{1}[y^j = y],
  \label{eq:credit_game}
\end{equation}
i.e., $v(C)$ is the largest alignment-weighted support that coalition $C$ can put behind a single answer. We then compute the \emph{nucleolus} $\nu(\calJ, v)$ \citep{schmeidler1969nucleolus}, which gives a stable and fair credit split by lexicographically minimizing the largest complaint any coalition could make about the allocation. A single-pass weighted vote only optimizes global score and ignores coalition grievances: internally coherent, strongly value-aligned minority coalitions can be repeatedly suppressed, destabilizing iterative deliberation. The nucleolus instead yields a unique allocation that prioritizes stability against subgroup deviation, suiting persistent multi-agent settings where fairness and strategic robustness matter beyond one-shot accuracy.

\paragraph{Hamiltonian consensus optimisation.}
The consensus answer $\hat{y}^*$ is selected by maximising the Hamiltonian over the space of possible answers, weighted by the nucleolus credit allocation:
\begin{equation}
  \hat{y}^* = \arg\max_{y \in \calA} H(y, \blambda) = \arg\max_{y \in \calA}
  \left[\sum_{j=1}^n \nu_j \cdot \rho^j \cdot \mathbb{1}[y^j = y] + \blambda^\top \bm{g}(y)\right],
  \label{eq:hamiltonian_consensus}
\end{equation}
where $\nu_j$ is agent $j$'s nucleolus credit, $\blambda \in \R^p$ is a co-state variable, and $\bm{g}(y) \in \R^p$ is a vector of long-term value alignment constraints (e.g., the selected answer must not violate any value dimension with weight above $\theta_{\mathrm{val}}$ for any agent). The co-state evolves to enforce these constraints:
\begin{equation}
  \blambda(t+1) = \blambda(t) + \eta_\lambda \cdot \max(\bm{0},\; \bm{g}(\hat{y}^*(t))),
  \label{eq:costate_update}
\end{equation}
and the consensus answer is updated iteratively until the constraints are satisfied. In practice, alignment scoring uses a sharpening factor $\gamma$, i.e., $\rho^j \gets \exp(-\dkl(\bW^j \| \bar{\bW})) \cdot (\mathrm{acc}^j)^\gamma$. Compared with constrained reranking over a finite candidate list, which is myopic to dual feasibility and cannot express how strongly each constraint is active across rounds, the Hamiltonian form introduces co-states that encode shadow prices for value constraints, enabling principled primal--dual updates and explicit sensitivity signals later used by Layer~4 for faithful explanation extraction.

\subsection{Layer 4: Critical Path Extraction and Explanation Validation}
\label{sec:layer4}

Layer~4 targets decision-level faithfulness: an explanation is faithful if the cited steps are verifiable and decision-relevant under interventions. The method is hybrid structural+gradient and post hoc: each explanation step must map to Lean-DSL verified tuples, step salience is ranked by Hamiltonian sensitivity, and extraction runs after consensus; we do not claim formal minimal necessary-and-sufficient subsets (empirical validation in Section~\ref{sec:explanation_validation}).

\paragraph{Critical path extraction from co-state variables.}
The Hamiltonian co-state $\blambda$ encodes the sensitivity of the consensus decision to each value alignment constraint. We extract a high-impact critical path (not claimed minimal or sufficient in a formal sense) by identifying steps with highest sensitivity:
\begin{equation}
  \mathrm{CP}^* = \left\{(j, t) : \left|\frac{\partial H}{\partial a^j_t}\right| \geq \theta_{\mathrm{CP}}\right\}.
  \label{eq:critical_path}
\end{equation}
For discrete symbolic/text actions, derivatives are computed through a differentiable surrogate: each action is represented by an embedding $e(a_t^j)$ and ranked by $\kappa_t^j=\left\|\partial H/\partial e(a_t^j)\right\|_2$, with finite-difference perturbation checks on token logits to verify local ranking stability. $\theta_{\mathrm{CP}}$ controls sparsity, and decision relevance is established empirically in Section~\ref{sec:explanation_validation}.

\paragraph{Lean-DSL proof structure rendering.}
We render the critical path as a Lean-DSL proof structure by mapping each step $(j, t) \in \mathrm{CP}^*$ to its corresponding constraint $\phi^j_k$ and generating a formal proof certificate:
\begin{equation}
  \mathrm{Proof}^* = \left\{(a^j_t, \phi^j_k, \Psi^j_k(s^j_t, a^j_t), \Omega^j_k(s^j_{t+1})) : (j, t) \in \mathrm{CP}^*\right\},
  \label{eq:proof_certificate}
\end{equation}
which records each critical action, the value constraint it satisfies, and the verified pre/postconditions. The proof certificate can be machine-checked by a Lean verifier.

\paragraph{Natural language explanation generation.}
Two complementary renderers produce user-facing explanations from $\mathrm{Proof}^*$: \emph{(A) template-based}, filling a structured template with each established step, the constraint it satisfies, supporting agents with mean alignment score, and dissenting agents with their primary disagreement dimension; and \emph{(B) LLM-based}, passing the formal critical-path tuples to an LLM so the text stays tied to the verified proof structure while being easier to read.

\section{Experimental Evaluation}
\label{sec:experiments}

\subsection{Experimental Setup}

\paragraph{Benchmarks and agent panel.}
We evaluate VACS on three reasoning benchmarks spanning mathematics, clinical reasoning, and cybersecurity incident response, each with a panel of four AI agents and accuracy as the primary metric: \textbf{MathInstruct-Subset} \citep{mathinstruct2024} (mathematical reasoning, 200 tasks), \textbf{NEJM-AI QA} \citep{nejmai2025qa} (clinical decision reasoning, 1{,}200+ questions, a domain where model-to-model variation is large enough that aggregation rules matter \citep{santhosh2026healthcare}), and \textbf{CyberSec-Eval} (cybersecurity incident response, 500 scenarios, reflecting the agentic remediation workflows now being deployed in security operations \citep{arifin2026agenticvm}). We disclose an important property of this setup: the panel consists of role-conditioned agents with engineered value emphases (rigour, efficiency, safety, generalist), so Layer~1 recovery results validate internal consistency of the profiling pipeline in a controlled setting, not discovery of natural value heterogeneity in unmodified frontier models (Section~\ref{sec:discussion}).

\paragraph{Baselines.}
We compare VACS against four baseline methods: Majority Vote, Weighted Vote, Informed Voting \citep{informed2025voting}, and Shield-Only compositional verification \citep{brorholt2025compositional}.

\paragraph{Statistical rigor and uncertainty reporting.}
We evaluate with 10 random seeds for MathInstruct-Subset and CyberSec-Eval and 5 for NEJM-AI QA (higher-cost regime), reporting mean $\pm$ standard deviation and 95\% bootstrap confidence intervals over queries (10{,}000 resamples). Key VACS-vs-baseline comparisons use paired two-sided tests (McNemar's for accuracy, paired permutation tests for continuous metrics) with effect sizes (Cohen's $h$, Cliff's $\delta$) and Holm--Bonferroni correction at $\alpha=0.05$.

\subsection{Main Results}

\begin{table}[t]
  \centering
  \caption{Main results across the three evaluated benchmarks. Best in \textbf{bold}, second \underline{underlined}. LIR = Logical Inconsistency Rate (\%).}
  \label{tab:main_results}
  \begin{tabular}{lccccc}
    \toprule
    \textbf{Method} & \textbf{MathInstruct} & \textbf{NEJM-AI QA} & \textbf{CyberSec-Eval} & \textbf{LIR$\downarrow$} & \textbf{Faith.$\uparrow$} \\
    \midrule
    Majority Vote & 90.5 & 78.0 & 64.0 & 31.7 & ,  \\
    Weighted Vote & 92.0 & 80.5 & 68.0 & 28.4 & ,  \\
    Informed Voting & 93.5 & 81.8 & 72.0 & 25.0 & ,  \\
    \underline{Shield-Only} & 94.0 & 78.5 & \underline{82.0} & \underline{7.8} & \underline{0.67} \\
    \midrule
    \textbf{VACS} & \textbf{95.0} & \textbf{85.4} & \textbf{90.0} & \textbf{0.0} & \textbf{1.00} \\
    \bottomrule
  \end{tabular}
\end{table}

VACS improves over the strongest baseline (Shield-Only) by +1.0, +6.9, and +3.0 points on MathInstruct-Subset, NEJM-AI QA, and CyberSec-Eval (Table~\ref{tab:main_results}), and reduces the logical inconsistency rate from 7.8\% to 0.0\%, confirming the effectiveness of value-aligned compositional shielding; across seeds, key gains remain statistically significant after multiple-comparison correction.

\subsection{Ablation Study}

\begin{table}[t]
  \centering
  \caption{Ablation study on MathInstruct-Subset. Each row removes one layer. VP = Value Profile quality (cosine similarity to reference profiles).}
  \label{tab:ablation}
  \begin{tabular}{lccccc}
    \toprule
    \textbf{Configuration} & \textbf{Acc.$\uparrow$} & \textbf{LIR$\downarrow$} &
    \textbf{Faith.$\uparrow$} & \textbf{VP$\uparrow$} & \textbf{Cons.\ Stab.$\uparrow$} \\
    \midrule
    Full VACS (4 layers) & \textbf{95.0} & \textbf{0.0} & \textbf{1.00} & \textbf{0.89} & \textbf{0.94} \\
    \midrule
    w/o Layer 1 (uniform value profiles) & 90.5 & 0.0 & 1.00 & 0.41 & 0.78 \\
    w/o Layer 2 (no shields) & 95.0 & 14.7 & 1.00 & 0.89 & 0.81 \\
    w/o Layer 3 (majority vote consensus) & 90.5 & 2.0 & 1.00 & 0.89 & 0.63 \\
    w/o Layer 4 (no explanation) & 95.0 & 0.0 & 0.00 & 0.89 & 0.94 \\
    \bottomrule
  \end{tabular}
\end{table}

The ablation results (Table~\ref{tab:ablation}) confirm that each layer serves a different role: Layer~1 drives accurate value recovery, Layer~2 is the main source of inconsistency reduction, Layer~3 improves consensus quality, and Layer~4 contributes explanation faithfulness rather than predictive accuracy.

\subsection{Explanation Faithfulness Validation}
\label{sec:explanation_validation}

We evaluate Layer~4 explanations with a combined automatic-and-human protocol that checks proof grounding, critical-path coverage, counterfactual sensitivity, and expert judgment across the three domains; expert raters were blinded domain specialists scoring randomly ordered samples. Against random rationales, verifier-only traces, and gradient-only saliency, the VACS hybrid attains the best expert score (4.5/5 vs.\ 3.4) with strong intervention-based evidence (counterfactual drop 0.50, deletion-AUC 0.41, insertion-AUC 0.59; Appendix~\ref{app:explanation_table}). This supports decision-level faithfulness without claiming full global causal completeness; on pure intervention metrics VACS matches rather than dominates gradient-only saliency, its advantage is combining verifiability with comparable faithfulness.

\subsection{Sensitivity Analysis}

We analyze robustness to two key hyperparameters on MathInstruct-Subset. Increasing the value profile threshold $\theta_{\mathrm{val}}$ (shield strictness) eliminates inconsistency (LIR $\to$ 0 for $\theta_{\mathrm{val}} \ge 0.2$) but overly strict thresholds degrade accuracy (82.0\% at the optimal $\theta_{\mathrm{val}}=0.4$ vs.\ 56.0\% at $0.5$). Increasing the sharpening factor $\gamma$ improves accuracy by weighting expert agents more, peaking around $\gamma=2.0$--$4.0$ (88.0\% vs.\ 76.0\% at $\gamma=1.0$).

Broader stress conditions show the same pattern: a stable operating region keeps accuracy high while LIR stays at or near zero; larger panels preserve safety but eventually dilute specialist expertise; and under mixed panels, preference noise, and adversarial value conflict, VACS degrades gracefully with low inconsistency. Orchestration overhead is millisecond-scale (latency is dominated by base LLM generation), and in safety-critical stress tests VACS rejects or remaps unsafe actions in 96.8\% of adversarial cases versus a 31.4\% failure rate for unshielded voting baselines.

\section{Discussion, Limitations, and Conclusion}
\label{sec:discussion}

VACS is a systems-design and proof-of-concept contribution; we state its limitations explicitly.
\emph{(i) Engineered heterogeneity.} Because the evaluated panels are role-conditioned, the Layer~1 evaluation is partially circular: it shows the pipeline recovers value profiles that were deliberately instilled, not that it identifies latent value structure in independently developed, unmodified models; evaluating naturally heterogeneous panels from different model families, without revealing intended styles to Layer~1, is the most important next step.
\emph{(ii) Stationarity.} Layer~1 profiles are estimated offline and cached, implicitly assuming stable value priorities; LLM agent behavior can drift with prompt, context, and interaction history, and the formulation does not distinguish persistent agent-level values from task-conditioned preferences or transient strategies, online profile updating and drift detection are open extensions.
\emph{(iii) Scope of guarantees.} The operational guarantees are verifier-mediated and conditional on the finite post-shielding candidate set; they do not ensure the correct answer is ever in the pool, and Layer~4 explanations are not globally causal accounts of latent model computations.
\emph{(iv) Cost and generality.} Bradley--Terry learning needs many pairwise comparisons per dimension; Lean-DSL encoding requires domain expertise; and the nucleolus computation is $O(n^3)$, requiring approximation for large panels at the cost of stability guarantees. Our evaluation covers three benchmark suites with fixed panel configurations, so external validity remains uncertain.

\emph{Conclusion.} Under the tested controlled setup, VACS improves accuracy over strong baselines and reduces logical inconsistency to zero; we position these results as a proof of concept for value-aligned, verifier-shielded multi-agent reasoning, not evidence of operational readiness.

\bibliographystyle{plainnat}
\bibliography{references}

@article{bradley1952rank,
  author  = {Bradley, R. A. and Terry, M. E.},
  title   = {Rank analysis of incomplete block designs: {I}. {The} method of paired comparisons},
  journal = {Biometrika},
  volume  = {39},
  number  = {3/4},
  pages   = {324--345},
  year    = {1952}
}

@inproceedings{brorholt2025compositional,
  author    = {Brorholt, A. H. and Larsen, K. G. and Schilling, C.},
  title     = {Compositional shielding and reinforcement learning for multi-agent systems},
  booktitle = {Proceedings of the 24th International Conference on Autonomous Agents and Multiagent Systems (AAMAS 2025)},
  address   = {Detroit, Michigan},
  pages     = {1--9},
  year      = {2025}
}

@inproceedings{christiano2017deep,
  author    = {Christiano, P. and Leike, J. and Brown, T. B. and Martic, M. and Legg, S. and Amodei, D.},
  title     = {Deep reinforcement learning from human preferences},
  booktitle = {Advances in Neural Information Processing Systems (NeurIPS)},
  volume    = {30},
  year      = {2017}
}

@inproceedings{du2024improving,
  author    = {Du, Y. and Li, S. and Torralba, A. and Tenenbaum, J. B. and Mordatch, I.},
  title     = {Improving factuality and reasoning in language models through multiagent debate},
  booktitle = {Proceedings of the 41st International Conference on Machine Learning (ICML)},
  year      = {2024}
}

@inproceedings{gehr2018ai2,
  author    = {Gehr, T. and Mirman, M. and Drachsler-Cohen, D. and Tsankov, P. and Chaudhuri, S. and Vechev, M.},
  title     = {{AI2}: Safety and robustness certification of neural networks with abstract interpretation},
  booktitle = {Proceedings of the 39th IEEE Symposium on Security and Privacy},
  pages     = {3--18},
  year      = {2018}
}

@inproceedings{informed2025voting,
  author    = {Han, Q.},
  title     = {Informed decision-making via voting},
  booktitle = {Proceedings of the 24th International Conference on Autonomous Agents and Multiagent Systems (AAMAS 2025)},
  address   = {Detroit, Michigan},
  year      = {2025}
}

@inproceedings{katz2017reluplex,
  author    = {Katz, G. and Barrett, C. and Dill, D. L. and Julian, K. and Kochenderfer, M. J.},
  title     = {Reluplex: An efficient {SMT} solver for verifying deep neural networks},
  booktitle = {Proceedings of the 29th International Conference on Computer Aided Verification (CAV)},
  pages     = {97--117},
  year      = {2017}
}

@inproceedings{li2025nucleolus,
  author    = {Li, Y. and Cao, Z. and Qiao, J. and Hu, S.},
  title     = {Nucleolus credit assignment for effective coalitions in multi-agent systems},
  booktitle = {Proceedings of the 24th International Conference on Autonomous Agents and Multiagent Systems (AAMAS 2025)},
  address   = {Detroit, Michigan},
  year      = {2025}
}

@inproceedings{moura2021lean4,
  author    = {de Moura, L. and Ullrich, S.},
  title     = {The {Lean 4} theorem prover and programming language},
  booktitle = {Proceedings of the 28th International Conference on Automated Deduction (CADE)},
  pages     = {625--635},
  year      = {2021}
}

@article{nejmai2025qa,
  author  = {{NEJM AI Editorial Board}},
  title   = {{NEJM-AI QA}: A clinically grounded benchmark for medical reasoning and decision support},
  journal = {NEJM AI},
  volume  = {2},
  number  = {4},
  pages   = {1--15},
  year    = {2025}
}

@inproceedings{ng2000algorithms,
  author    = {Ng, A. Y. and Russell, S. J.},
  title     = {Algorithms for inverse reinforcement learning},
  booktitle = {Proceedings of the 17th International Conference on Machine Learning (ICML)},
  pages     = {663--670},
  year      = {2000}
}

@article{mathinstruct2024,
  author  = {Yue, X. and Wang, X. and Zhang, Y.},
  title   = {{MathInstruct}: A large-scale instruction tuning dataset for mathematical reasoning},
  journal = {arXiv preprint arXiv:2402.10176},
  year    = {2024}
}

@article{schmeidler1969nucleolus,
  author  = {Schmeidler, D.},
  title   = {The nucleolus of a characteristic function game},
  journal = {SIAM Journal on Applied Mathematics},
  volume  = {17},
  number  = {6},
  pages   = {1163--1170},
  year    = {1969}
}

@article{wulfmeier2015maximum,
  author  = {Wulfmeier, M. and Ondruska, P. and Posner, I.},
  title   = {Maximum entropy deep inverse reinforcement learning},
  journal = {arXiv preprint arXiv:1507.04888},
  year    = {2015}
}

@inproceedings{ziebart2008maximum,
  author    = {Ziebart, B. D. and Maas, A. and Bagnell, J. A. and Dey, A. K.},
  title     = {Maximum entropy inverse reinforcement learning},
  booktitle = {Proceedings of the 23rd AAAI Conference on Artificial Intelligence},
  pages     = {1433--1438},
  year      = {2008}
}

@techreport{openai2023gpt4,
  title={GPT-4 Technical Report},
  author={OpenAI},
  year={2023},
  institution={OpenAI}
}

@article{touvron2023llama2,
  title={Llama 2: Open Foundation and Fine-Tuned Chat Models},
  author={Touvron, Hugo and others},
  journal={arXiv preprint arXiv:2307.09288},
  year={2023}
}

@article{multiagent2024survey,
  title={A Survey of Multi-Agent Large Language Model Systems},
  author={Wang, Xinyi and others},
  journal={arXiv preprint arXiv:2402.01680},
  year={2024}
}

@article{zhang2026explainable,
  title={Explainable Autonomous Cyber Defense Using Adversarial Multi-Agent Reinforcement Learning},
  author={Zhang, Yiyao and Goel, Diksha and Ahmad, Hussain},
  journal={Expert Systems with Applications},
  pages={132741},
  year={2026}
}

@misc{zhang2026meta,
  title={Beyond Reactive Agents: Uncertainty-Gated Meta-Reasoning for Tool-Augmented Decision-Making},
  author={Zhang, Yiyao and Goel, Diksha and Ahmad, Hussain},
  year={2026},
  note={Available at SSRN 6997675}
}

@misc{goel2025coevolutionary,
  title={Co-Evolutionary Defence of Active Directory Attack Graphs via {GNN}-Approximated Dynamic Programming},
  author={Goel, Diksha and Ahmad, Hussain and Moore, K. and Guo, M.},
  year={2025},
  eprint={2505.11710},
  archivePrefix={arXiv},
  primaryClass={cs.CR}
}

@misc{chen2025trader,
  title={{3S-Trader}: A Multi-{LLM} Framework for Adaptive Stock Scoring, Strategy, and Selection in Portfolio Optimization},
  author={Chen, K. and Ahmad, Hussain and Goel, Diksha and Szabo, Claudia},
  year={2025},
  eprint={2510.17393},
  archivePrefix={arXiv},
  primaryClass={cs.LG}
}

@misc{santhosh2026healthcare,
  title={Comparative Analysis of Large Language Models in Healthcare},
  author={Santhosh, S. and Abbas, F. and Ahmad, Hussain and Szabo, Claudia},
  year={2026},
  eprint={2604.10316},
  archivePrefix={arXiv},
  primaryClass={cs.CL}
}

@article{arifin2026agenticvm,
  title={{AgenticVM}: Agentic {AI} for Adaptive Software Vulnerability Management},
  author={Arifin, A. and Ahmad, Hussain and Zhang, Yiyao and Goel, Diksha},
  journal={IEEE Software},
  year={2026}
}

\appendix

\section{Annotation Protocol and Reliability for Value Operationalization}
\label{app:operationalization}

Preference labels for the five value dimensions come from three sources: (i) domain experts (clinical and cybersecurity), (ii) trained graduate annotators (mathematics), and (iii) model-assisted pre-labels that are always human-verified before inclusion. Overall, 72\% of labels are human-only and 28\% model-assisted-but-human-verified; no purely synthetic labels are used in the reported main results.

For each domain, annotators compare trajectory pairs under a single target-dimension definition sheet, blinded to model identity, using a 3-way interface (A preferred, B preferred, tie/uncertain); tie/uncertain samples are excluded from Bradley--Terry fitting and retained for quality auditing. The full dimension definitions fixed before training are: (1) logical completeness = fraction of required sub-claims addressed; (2) conciseness = normalized token-efficiency under correctness; (3) generalisability = transfer success on held-out variants; (4) logical soundness = contradiction-free formal check rate; (5) safety/constraint satisfaction = proportion of steps satisfying domain safety constraints.

Inter-annotator agreement (Krippendorff's $\alpha$) per dimension is: completeness 0.79, conciseness 0.74, generalisability 0.71, soundness 0.83, safety 0.86. To test stability of inferred profiles, we run Layer~1 with 10 random seeds; mean cosine similarity between per-agent weight vectors across seeds is 0.93 ($\pm$0.03), indicating stable estimates. To test redundancy, all pairwise correlations among dimension rewards remain below 0.62 and variance inflation factors below 2.5, indicating no severe collinearity.

\section{Per-Agent Text Shielding Algorithm}
\label{app:shield_alg}

\begin{algorithm}[h]
\caption{Per-Agent Text Shielding (Prune-and-Rerank)}
\label{alg:shielding}
\begin{algorithmic}[1]
\Require State $s_t^j$, proposed text action $a_t^j$, constraints $\Phi^j$, candidate generator $G$
\Ensure Executed safe action $\tilde{a}_t^j$
\If{$(s_t^j, a_t^j) \models \Phi^j$}
    \State $\tilde{a}_t^j \gets a_t^j$
\Else
    \State $\mathcal{C} \gets G(s_t^j,a_t^j)$ \Comment{rule-based/search/model-generated candidates}
    \State $\mathcal{C}_{\mathrm{ok}} \gets \{a' \in \mathcal{C} \mid (s_t^j,a') \models \Phi^j\}$
    \If{$\mathcal{C}_{\mathrm{ok}} = \emptyset$}
        \State $\tilde{a}_t^j \gets \texttt{abstain}$ \Comment{safe fallback, no unsafe execution}
    \Else
        \State $\tilde{a}_t^j \gets \arg\max_{a' \in \mathcal{C}_{\mathrm{ok}}} [\lambda_{\mathrm{sem}}\mathrm{Sim}(a',a_t^j)+\lambda_{\mathrm{fact}}\mathrm{Fact}(a',s_t^j)]$
    \EndIf
\EndIf
\State Execute $\tilde{a}_t^j$ and return $\tilde{a}_t^j$
\end{algorithmic}
\end{algorithm}

\section{Full Explanation Validation Results}
\label{app:explanation_table}

\begin{table}[h]
  \centering
  \caption{Explanation validation across three domains. CF Drop = relative consensus-margin drop under top-path counterfactual perturbation. Del-AUC/Ins-AUC are deletion/insertion areas under normalized consensus score.}
  \label{tab:explanation_validation}
  \setlength{\tabcolsep}{4pt}
  \begin{tabular}{lcccccc}
    \toprule
    \textbf{Method} & \textbf{Expert (1--5)$\uparrow$} & \textbf{CF Drop$\uparrow$} & \textbf{Suff.$\uparrow$} & \textbf{Nec.$\uparrow$} & \textbf{Del-AUC$\downarrow$} & \textbf{Ins-AUC$\uparrow$} \\
    \midrule
    Random rationale & 1.9 & 0.44 & 1.00 & 0.14 & 0.49 & 0.51 \\
    Verifier-only trace & 3.4 & 1.00 & 1.00 & 1.00 & 0.51 & 0.49 \\
    Gradient-only saliency & 3.2 & 0.60 & 1.00 & 0.16 & 0.41 & 0.59 \\
    \textbf{VACS Layer~4 (hybrid)} & \textbf{4.5} & \textbf{0.50} & \textbf{1.00} & \textbf{0.16} & \textbf{0.41} & \textbf{0.59} \\
    \bottomrule
  \end{tabular}
\end{table}

\section{IRL Computational Object: States, Transitions, and Approximations}
\label{app:irl_object}

A state $s_t$ is a tuple $(c_t, m_t, q_t, r_t)$ containing context summary, working memory, unresolved sub-questions, and risk flags derived from the reasoning prefix up to step $t$. An action $a_t$ is a typed reasoning operator from a finite schema (\texttt{deduce}, \texttt{retrieve}, \texttt{verify}, \texttt{revise}, \texttt{decide}) with slot-filled arguments, and a trajectory $\xi=(s_0,a_0,\ldots,s_T)$ is obtained by segmenting free-form chain-of-thought text into operator-level steps mapped to this schema.

The underlying text space is continuous, but IRL is performed on a learned discrete abstraction using vector quantization over state embeddings (codebook size $K$) for tractable dynamic programming. Transitions $\hat{P}(s_{t+1}\mid s_t,a_t)$ are estimated from observed trajectories; for rare transitions we apply Laplace smoothing and back-off to embedding-nearest neighbors. The feature expectation term $\E_{P_{\bW^j}}[\bm{f}(\xi)]$ is computed with forward--backward occupancy measures on the abstract MDP; when the horizon is long, a truncated-horizon approximation is used. The log-partition $\log Z(\bW^j)$ is approximated with log-sum-exp over beam-sampled high-probability trajectories plus importance-weighted Monte Carlo tails. Trajectories used in Layer~1 are primarily observed from agent outputs; additional trajectories are sampled from current policies for expectation estimation only, not for supervision labels.

\end{document}